\documentclass[letterpaper, 10 pt, conference]{ieeeconf} 
\usepackage{cite}
\usepackage{amsmath,amssymb,amsfonts}
\usepackage{algorithm}
\usepackage{algpseudocode}
\usepackage{xcolor}
\usepackage{graphicx}
\usepackage{hyperref}
\usepackage{cases}
\hypersetup{
    colorlinks=true,
    linkcolor=blue,
    filecolor=magenta,      
    urlcolor=cyan,
}
\usepackage{varwidth}
\usepackage{microtype}
\usepackage{subcaption}
\usepackage{booktabs} 
\usepackage{multirow}
\usepackage{pifont}

\usepackage{colortbl}
\definecolor{LightCyan}{rgb}{0.88,1,1}
\definecolor{Gray}{gray}{0.9}
\usepackage{ragged2e}

\usepackage[latin1]{inputenc}

\IEEEoverridecommandlockouts                              

\title{\LARGE \bf
IROS 2019 Lifelong Robotic Vision Challenge \\ Lifelong Object Recognition Report}

\justifying\author{
Qi She, Fan Feng, Qi Liu, Rosa H. M. Chan, Xinyue Hao, Chuanlin Lan, Qihan Yang, \\ Vincenzo Lomonaco, German I. Parisi, Heechul Bae, Eoin Brophy, Baoquan Chen, Gabriele Graffieti, \\ Vidit Goel, Hyonyoung Han, Sathursan Kanagarajah, Somesh Kumar, Siew-Kei Lam, Tin Lun Lam, \\
Liang Ma, Davide Maltoni,
Lorenzo Pellegrini, Duvindu Piyasena, Shiliang Pu, Debdoot Sheet, \\ Soonyong Song,
Youngsung Son, Zhengwei Wang,  Tom\'as E. Ward, Jianwen Wu, Meiqing Wu, Di Xie, \\ Yangsheng Xu, Lin Yang, Qiaoyong Zhong, Liguang Zhou
\thanks{Author list is in alphabetical order. F. Feng, Q. Liu, C. Lan, X. Hao, Q. Yang, Q. She, V. Lomonaco, G.I. Parisi and R.H.M. Chan prepared the report. Other co-authors were competition finalists.}
\thanks{R.H.M Chan, F. Feng, Q. Liu, Q. Yang are with the Department of Electrical Engineering, City University of Hong Kong, Hong Kong, China.}%
\thanks{X. Hao is with the Department of Electronic Engineering, Tsinghua University and Beijing University of Posts and Telecommunications, Beijing, China.}%
\thanks{C. Lan is with the School of Electronic Information, Wuhan University, Wuhan, China.}
\thanks{V. Lomonaco is with ContinualAI Research and University of Bologna, Italy}
\thanks{G.I. Parisi is with ContinualAI Research and University of Hamburg, Germany}
\thanks{Q. She is with Intel Labs and ContinualAI Research}
\thanks{Corresponding author: qi.she@intel.com}
}

\begin{document}
\maketitle
\thispagestyle{empty}
\pagestyle{empty}

\begin{abstract}
This report summarizes IROS 2019-Lifelong Robotic Vision Competition (Lifelong Object Recognition Challenge) with methods and results from the top $8$ finalists (out of over~$150$ teams). The competition dataset (L)ifel(O)ng (R)obotic V(IS)ion (OpenLORIS) - Object Recognition (OpenLORIS-object) is designed for driving lifelong/continual learning research and application in robotic vision domain, with everyday objects in home, office, campus, and mall scenarios. The dataset explicitly quantifies the variants of illumination, object occlusion, object size, camera-object distance/angles, and clutter information. Rules are designed to quantify the learning capability of the robotic vision system when faced with the objects appearing in the dynamic environments in the contest.  Individual reports, dataset information, rules, and released source code can be found at the \href{https://lifelong-robotic-vision.github.io/competition/}{\underline{project homepage}}.
\end{abstract}

\section{INTRODUCTION}
\begin{figure}[!ht]
    \centering
    \includegraphics[width=\linewidth]{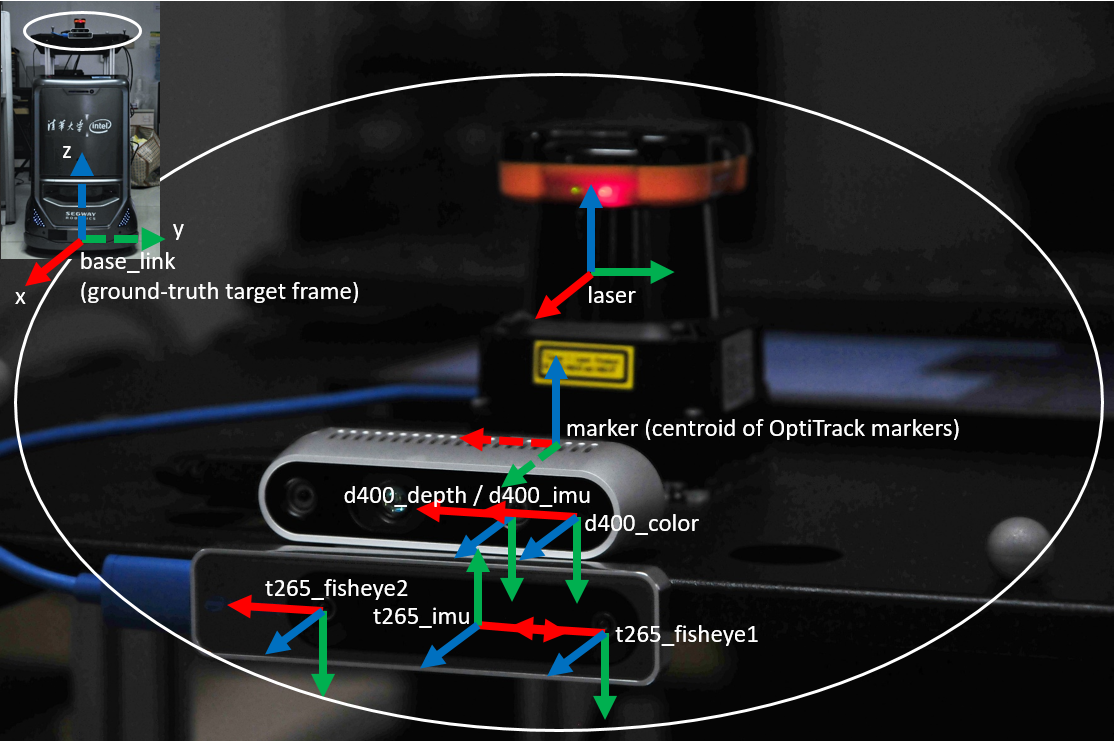}
    \caption{OpenLORIS robotic platform (left) mounted with multiple sensors (right). In OpenLORIS-Object dataset, the RGB-D data is collected from the depth camera.}
    \label{fig:robot}
\end{figure}

Humans have the remarkable ability to learn continuously from the external environment and the inner experience. One of the grand goals of robots is also building an artificial ``lifelong learning" agent that can shape a cultivated understanding of the world from the current scene and their previous knowledge via an autonomous lifelong development. It is challenging for the robot learning process to retain earlier knowledge when they encounter new tasks or information. Recent advances in computer vision and deep learning methods have been very impressive due to large-scale datasets, such as ImageNet~\cite{deng2009imagenet} and COCO~\cite{lin2014microsoft}. However, robotic vision poses unique new challenges for applying visual algorithms developed from these computer vision datasets because they implicitly assume a fixed set of categories and time-invariant task distributions~\cite{8911341}. Semantic concepts change dynamically over time~\cite{she2019neural, she2018stochastic, she2018reduced}. Thus, sizeable robotic vision datasets collected from real-time changing environments for accelerating the research and evaluation of robotic vision algorithms are crucial. For bridging the gap between robotic vision and stationary computer vision fields, we utilize a real robot mounted with multiple-high-resolution sensors (e.g., monocular/RGB-D from RealSense D435i, dual fisheye images from RealSense T265, LiDAR,, see Fig.~\ref{fig:robot}) to actively collect the data from the real-world objects in several kinds of typical scenarios, like homes, offices,campus, and malls. 

Lifelong learning approaches can be divided into 1) methods that retrain the whole network via regularizing the model parameters learned from previous tasks, e.g., Learning without Forgetting (LwF)~\cite{li2017learning}, Elastic Weight Consolidation (EWC)~\cite{kirkpatrick2017overcoming} and Synaptic Intelligence (SI)~\cite{zenke2017continual}; 2) methods that dynamically expand/adjust the network architecture if learning new tasks, e.g., Context-dependent Gating (XdG)~\cite{masse2018alleviating} and Dynamic Expandable Network (DEN)~\cite{yoon2017lifelong}; 3) rehearsal approaches gather all methods that save raw samples as memory of past tasks. These samples are used to maintain knowledge about the past in the model and then replayed with samples drawn from the new task when training the model, e.g., Incremental Classifier and Representation Learning (ICaRL)~\cite{rebuffi2017icarl}; and generative replay approaches train generative models on the data distribution~\cite{wang2019generative, wang2019neuroscore, wang2020neuro}, and they are able to afterward sample data from experience when learning new data, e.g., Deep Generative Replay (DGR)~\cite{shin2017continual}, DGR with dual memory~\cite{kamra2017deep} and feedback~\cite{van2018generative}.

This report summarizes IROS 2019-Lifelong Robotic Vision Competition (Lifelong Object Recognition challenge) with dataset, rules, methods and results from the top $8$ finalists (out of over $150$ teams). Individual reports, dataset information, rules, and released source codes can be found at the \href{https://lifelong-robotic-vision.github.io/competition/}{\underline{competition homepage}}.

\section{IROS 2019 Lifelong Robotic Vision - Object Recognition Challenge}
This challenge aimed to explore how to leverage the knowledge learned from previous tasks that could generalize to new task effectively, and also how to efficiently memorize of previously learned tasks. The work pathed the way for robots to behave like  humans in terms of knowledge transfer, association, and combination capabilities.

To our best knowledge, the provided lifelong object recognition dataset OpenLORIS-Object-v$1$~\cite{she2019openlorisobject} is the first one that explicitly indicates the task difficulty under the incremental setting, which is able to foster the lifelong/continual/incremental learning in a supervised/semi-supervised manner. Different from previous instance/class-incremental task, the difficulty-incremental learning is to test the model's capability over continuous learning when faced with multiple environmental factors, such as illumination, occlusion, camera-object distances/angles, clutter, and context information in both low and high dynamic scenes.

\subsection{OpenLORIS-Object Dataset}
IROS 2019 competition provided the $1^{st}$ version of OpenLORIS-Object dataset for the participants. Note that our dataset has been updated with twice the size in content available at the \href{https://lifelong-robotic-vision.github.io/dataset/Data_Object-Recognition}{\underline{project homepage}} with detailed information,visualization, downloading instructions and benchmarks on SOTA lifelong learning methods~\cite{she2019openlorisobject}.       

We included the common challenges that the robot is usually faced with, such as illumination, occlusion, camera-object distance, etc. Furthermore, we explicitly decompose these factors from real-life environments and have quantified their difficulty levels. In summary, to better understand which characteristics of robotic data negatively influence the results of the lifelong object recognition, we independently consider: 1) illumination, 2) occlusion, 3) object size, 4) camera-object distance, 5) camera-object angle, and 6) clutter. 
\begin{itemize}
    \item[1).]\textbf{Illumination}. The illumination can vary significantly across time, e.g., day and night. We repeat the data collection under weak, normal, and strong lighting conditions, respectively. The task becomes challenging with lights to be very weak.
    \item[2).]\textbf{Occlusion}. Occlusion happens when a part of an object is hidden by other objects, or only a portion of the object is visible in the field of view. Occlusion significantly increases the difficulty for recognition.
    \item[3).]\textbf{Object size}. Small-size objects make the task challenging, like dry batteries or glue sticks.
    \item[4).]\textbf{Camera-object distance}. It affects actual pixels of the objects in the image.
    \item[5).]\textbf{Camera-object angle}. The angles between the cameras and objects affect the attributes detected from the object.
    \item[6).]\textbf{Clutter}. The presence of other objects in the vicinity of the considered object may interfere with the classification task.
\end{itemize}

\begin{table*}[t!]
    \centering
    \begin{tabular}{c|c|c|c|c|c|c|c}
    \toprule
         Level & Illumination & Occlusion (percentage) & Object Pixel Size (pixels) & Clutter & Context & \#Classes & \#Instances \\ \hline
         1   &  Strong & $0\%$ & $> 200 \times 200$ & Simple & \multirow{3}{*}{Home/office/mall} & \multirow{3}{*}{19} & \multirow{3}{*}{69}  \\ 
         2  &  Normal      & $25\%$ & $30 \times 30-200\times 200$ & Normal & &\\ 
         3 & Weak    & $50\%$      &   $< 30 \times 30$ & Complex & &\\
    \bottomrule
    \end{tabular}
    \caption{Details of each $3$ levels for $4$ real-life robotic vision challenges.}
    \label{Task_level}
\end{table*}

The version of OpenLORIS-Object for this competition is a collection of $69$ instances, including $19$ categories daily necessities objects under $7$ scenes. For each instance, a $17$ seconds video (at $30$ fps) has been recorded with a depth camera delivering $~500$ RGB-D frames (with $260$ distinguishable object views picked and provided in the dataset). $4$ environmental factors, each has $3$ level changes, are considered explicitly, including illumination variants during recording, occlusion percentage of the objects, object pixel size in each frame, and the clutter of the scene. Note that the variables of 3) object size and 4) camera-object distance are combined together because in the real-world scenarios, it is hard to distinguish the effects of these two factors brought to the actual data collected from the mobile robots, but we can identify their joint effects on the actual pixel sizes of the objects in the frames roughly. The variable 5) is considered as different recorded views of the objects. The defined three difficulty levels for each factor are shown in Table.~\ref{Task_level} (totally we have $12$ levels w.r.t. the environment factors across all instances). The levels $1$, $2$, and $3$ are ranked with increasing difficulties. 

\begin{figure*}[t!]
\centering
  \includegraphics[width=0.93\linewidth]{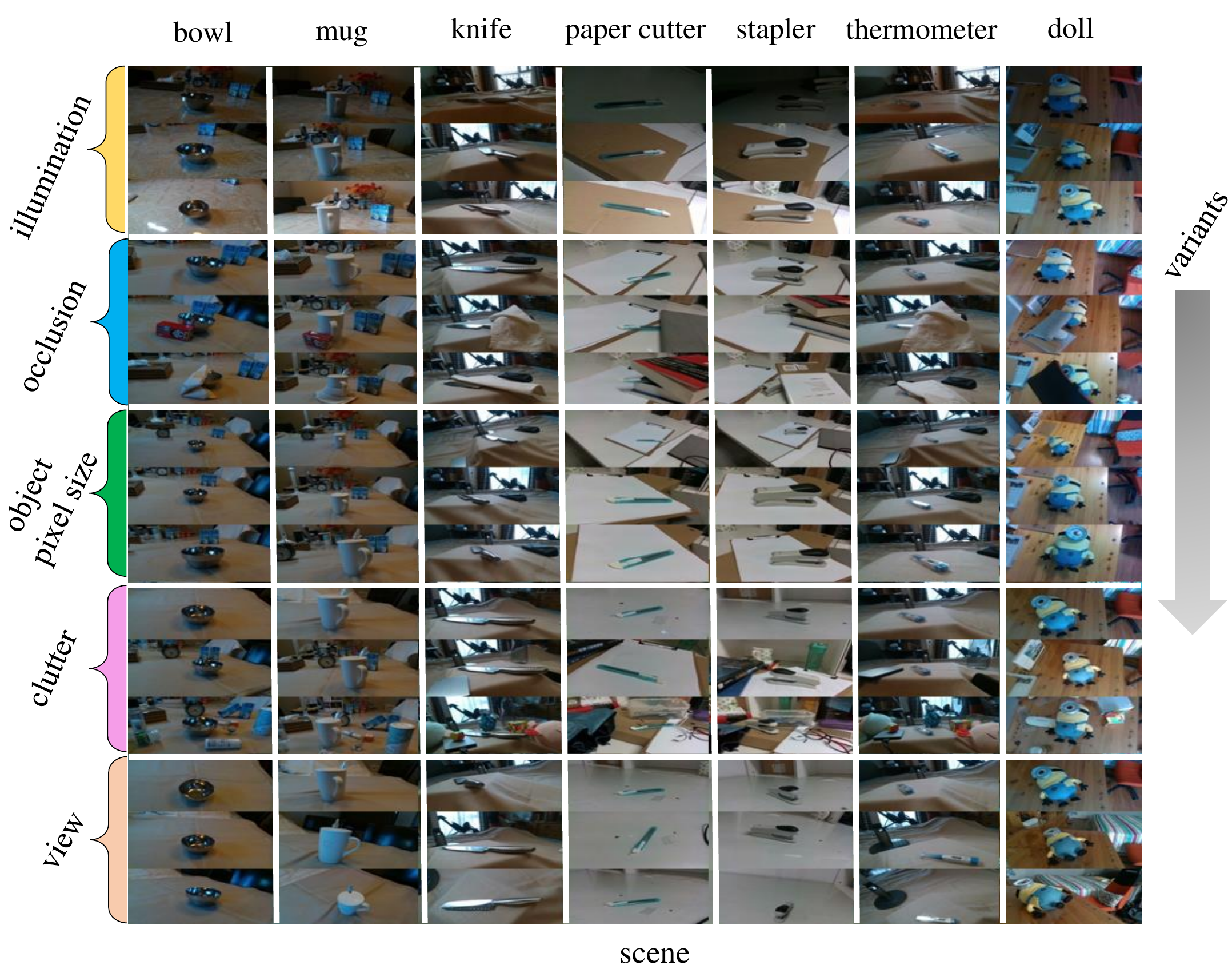}
  \caption{Picked samples of the objects from $7$ scenes (column) under multiple level environment conditions (row). The variants from top to bottom are illumination (weak, normal, and strong); occlusion ($0\%$, $25\%$, and $50\%$); object pixel size ($<30 \times 30$, $30 \times 30 - 200 \times 200$, and $ > 200 \times 200$); clutter (simple, normal and complex); and multi-views of the objects. (Note that we use different views as training samples of each difficulty level in each factor). \vspace{-2mm}
  }
  \label{fig:object}
\end{figure*}

For each instance at each level, we provided $260$ samples, both have RGB and depth images. Thus, the total images provided is around $2$ (RGB and depth) $\times 260$ (samples per instance)$\times 69$ (instances) $\times 4$ (factors per level) $\times 3$ (difficulty levels) = $430,560$ images. Also, we have provided bounding boxes and masks for each RGB image with Labelme~\cite{russell2008labelme}. The size of images under illumination, occlusion and object pixel size factors is 424$\times$240 pixels, and the size of images under object pixel size factor are 424$\times$240, 320$\times$180, 1280$\times$720 pixels (for $3$ difficulty levels). Picked samples have been shown in Fig.~\ref{fig:object}. 

\subsection{Challenge Phases and Evaluation Rules}
We held $2$ phases for the challenge. The preliminary contest we provided $9$ batches of datasets which contain different factors and difficulty levels, for each batch, we have train/validation/test data splits. The core of this incremental learning setting is, we need the first train on the first batch of the dataset, and then $2^{nd}$ batch, $3^{rd}$ batch, until the $9^{th}$ batch, and then use the final model to obtain the test accuracy of all encounter tasks (batches). The training/validation datasets can only be accessed during the model optimizations. We held the evluation platform on \href{https://codalab.lri.fr/competitions/581}{\underline{Codalab}}. There had been over over $150$ participants during the preliminary contest and we chose $8$ teams with higher testing accurries over all testing batches as our finalists.

For the final round, different from standard computer vision challenge~\cite{deng2009imagenet, lin2014microsoft}, not only the overall accuracy on all tasks was evaluated but also the model efficiency, including model size, memory cost, and replay size (the number of old task samples used for learning new tasks, smaller is better) were considered. Meanwhile, instead of directly asking the participants to submit the prediction results on the test dataset as standard deep learning challenges~\cite{deng2009imagenet, lin2014microsoft}, the organizers received either source codes or binary codes to evaluate their whole lifelong learning process to make fair comparison. The finalists' methods were tested by the organizers on Intel Core i9 CPU and 1Nvidia RTX 1080 Ti GPU. For final round dataset, we randomly shuffled the dataset with multiple factors. Data is split up to $12$ batches/tasks and each batch/task samples are from one subdirectories (there are $12$ subdirectories in total, $4$ factors $\times$ $3$ level/factor). Each batch includes 69 instances from $7$ scenes, about $21520$ test samples, $21,520$ validation samples and $172,200$ training samples. The metrics and corresponding grading weights are shown in Table~\ref{tbl:metrics}. As can be seen, we also provided a bonus test set which is recorded in under different context background with some deformation. The adaptation on this bonus testing data is a challenging task for our task.

\begin{table*}[!ht]
\centering
\caption{Metrics and grading criteria for final round}
\label{tbl:metrics}
\begin{tabular}{c|c|c|c|c|c|c}
\toprule
Metric & Accuracy & Model Size & Inference Time & Replay Size & Oral Presentation & Accuracy on Bonus Dataset\\ \hline

Weight & $50\%$       & $8\%$   &   $8\%$   & $8\%$   & $10\%$ & $16\%$     \\ \bottomrule
\end{tabular}
\end{table*}

\subsection{Challenge Results}
From more than $150$ registered participants, $8$ teams entered in the final phase and submitted results, codes, posters, slides and abstract papers (\href{https://lifelong-robotic-vision.github.io/competition/}{\underline{available here}}). Table~\ref{tbl:results} reports the details of all metrics (except oral presentation) for each team. 

\textbf{Architectures and main ideas: } All the proposed methods use end-to-end
deep learning models and employ the GPU(s) for training. For lifelong learning strategies: $5$ teams applied regularization methods, $2$ teams utilized knowledge distillation methods and $1$ team used network expansion method. $4$ teams applied resampling mechanism to alleviate catastrophic forgetting. Meanwhile, some other computer vision methods including saliency map, Single Shot multi-box Detection (SSD), data augmentation are also utilized in their solutions.  

\newcommand{\tabincell}[2]{\begin{tabular}{@{}#1@{}}#2\end{tabular}}  

\begin{table*}[!ht]
\centering
\caption{IROS 2019 Lifelong Robotic Vision Challenge final results.}
\label{tbl:results}
\begin{tabular}{cccccc}
\toprule
Teams & \tabincell{c}{Final \\Acc. (\%)} & \tabincell{c}{Model \\Size (MB)} & \tabincell{c}{Inference \\ time (s)} & \tabincell{c}{Replay \\ Size (\#sample)} & \tabincell{c}{Bonus-set\\ Acc. (\%)}\\ \hline
HIK\_ILG & $96.86$       & $16.30$   &   $25.42$   & $\underline{\bf{0}}$   & $\underline{\bf{21.86}}$     \\ \hline
Unibo & $97.68$       & $\underline{\bf{5.900}}$   &   $\underline{\bf{22.41}}$   & $1,500$   & $8.500$     \\ \hline
Guiness & $72.90$       & $9.400$   &   $346.0$   & $\underline{\bf{0}}$   & $10.96$      \\ \hline
Neverforget & $92.93$       & $342.9$   &   $467.1$   & $\underline{\bf{0}}$   & $1.520$      \\ \hline
SDU\_BFA\_PKU & $\underline{\bf{99.56}}$       & $171.4$   &   $2,444$   & $28,500$   & $19.54$     \\ \hline
Vidit98 & $96.16$       & $9.400$   &   $112.2$   & $1,300$   & $1.390$      \\ \hline
HYDRA-DI-ETRI & $10.42$       & $13.40$   &   $1,323$   & $21,312$   & $7.100$      \\ \hline
NTU\_LL & $93.56$       & $467.1$   &   $4,213$   & $\underline{\bf{0}}$   & $2.100$      \\ 

\bottomrule
\end{tabular}
\end{table*}

\section{Challenge Methods and Teams}
\subsection{\textbf{HIK\_ILG Team}} 
\begin{figure}[htbp]
    \centering
    \includegraphics[width=0.45\textwidth]{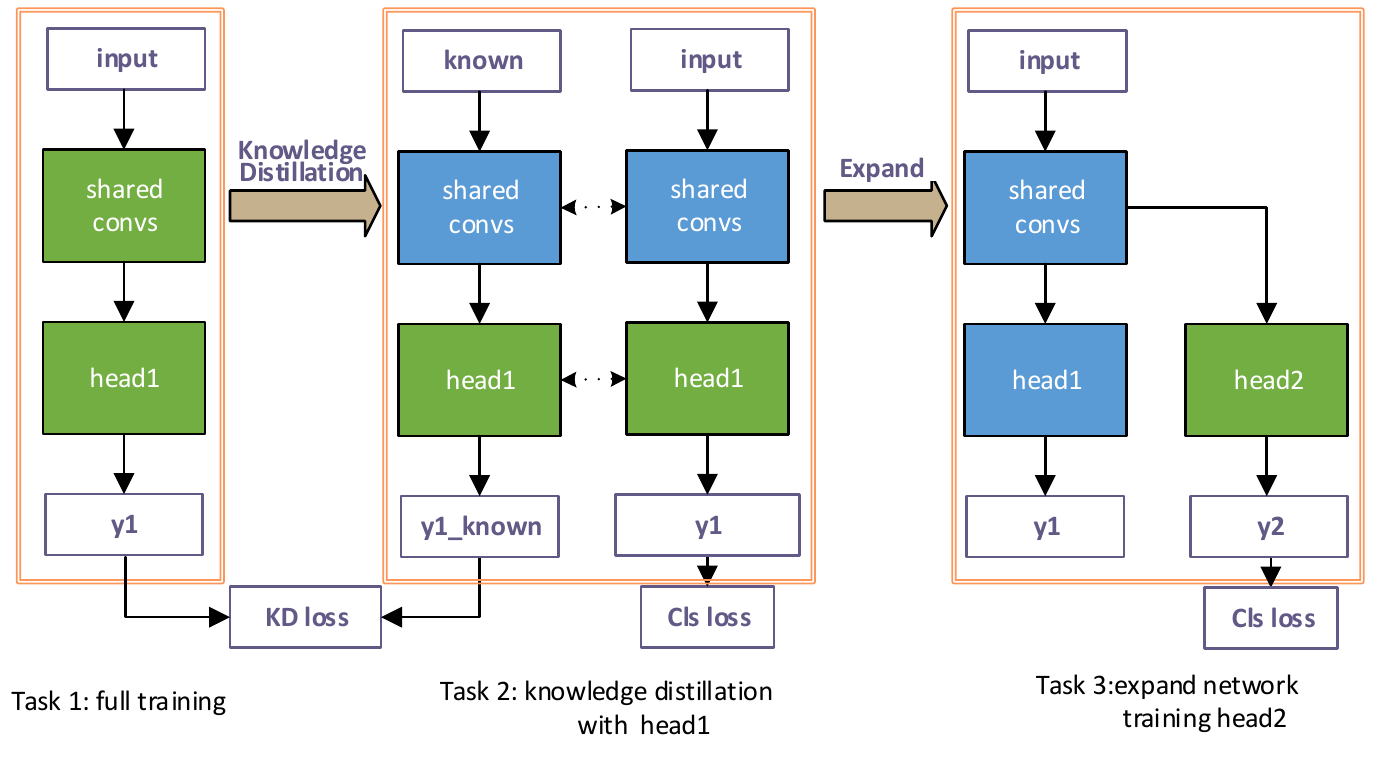}
    \caption{The architecture of proposed dynamic neural network by HIK\_ILG Team.}
    \label{fig:hik_arch}
\end{figure}

The team developed the dynamic neural network, which was comprised of two parts: dynamic network expansion for data across dissimilar domains and knowledge distillation for data in similar domains (See Figure~\ref{fig:hik_arch}). They froze the shared convolutional layers and trained new heads for new tasks. The domain gap was determined by measuring the accuracy of the previous model before training on current task. In order to increase the generalization ability of the trained model, they used ImageNet pre-trained model for the shared convolutional layers, and took more data augmentation and more batches to train head1 for base model. Without using previous data, they discovered known instances in current task by a single forward pass via previous model. Those correctly classified were treated as known samples. They used these samples for knowledge distillation. They utilized the best head over multiple heads for distillation, which is verified by experimental results. 

\subsection{\textbf{Unibo Team}}
\begin{figure}[!h]
\centerline{\includegraphics[scale=0.8]{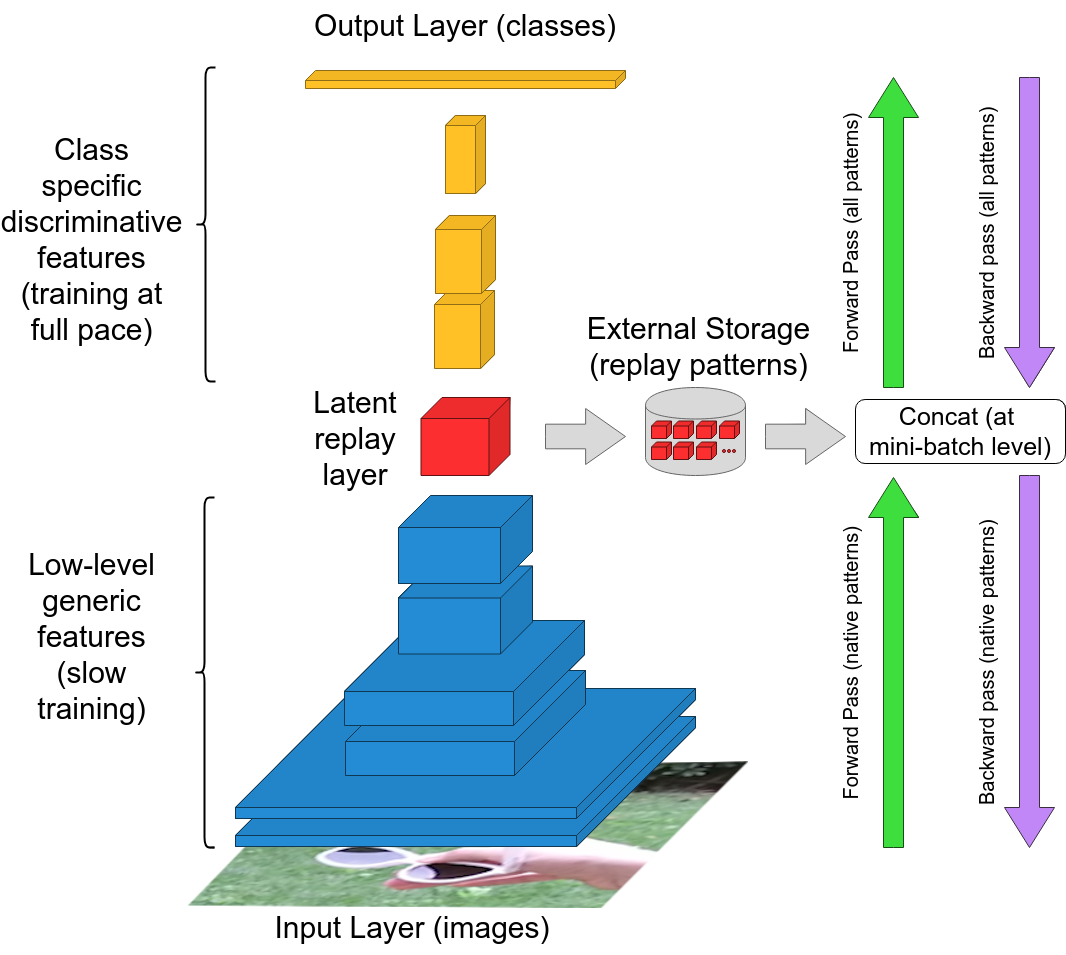}}
\caption{Architectural diagram of latent rehearsal in Unibo Team.}
\label{latent}
\end{figure}

The team proposed a new Continual Learning approach based on latent rehearsal, namely the replay of latent
neural network activation instead of raw images at the input
level. The algorithm can be deployed on the edge with low latency. With latent rehearsal (see Figure~\ref{latent}) they denoted an approach
where instead of maintaining in the external memory copies
of input patterns in the form of raw data, they stored the pattern
activation at a given level (denoted as latent rehearsal layer). The algorithm can be summarized as follow: 
1) Take $n$ patterns from the current batch;
2) Forward them through the network until the rehearsal layer;
3) Select $k$ patterns from the rehearsal memory;
4) Concat the original and the replay patterns;
5) Forward all the patterns through  the rest of the network;
6) Backpropagate the loss only until the rehearsal layer.

The specific design they utilized with was AR1*, AR1*free
and LwF CL approaches over a MobileNet-v1 and
MobileNet-v2~\cite{sandler2018mobilenetv2, nguyen2019contcap, maltoni2019continuous, howard2017mobilenets}. Meanwhile, they opted for simplicity and the trivial
rehearsal approach summarized in Algorithm~\ref{al:latent} is used for memory management.

\begin{algorithm}[!ht]
    \caption{Pseudo-code explaining how the external memory $M$  is populated across the training batches.}\label{al:latent}
    \begin{algorithmic}
        \item[\textbf{Require: }$M=\emptyset$]
        \item[\textbf{Require: }$M_{size}=\text{number of patterns to be stored in }M$]
        \item[\textbf{For each }training batch $B_i$ \textbf{do}]
        \State train the model on shuffled $B_i \cup M$
        \State $h=M_{size}/i$
        \State $R_{add}=\text{Random sampling }h \text{patterns from }B_i$
        \State $R_{replace}=
                \begin{cases}
                \text{Sample } h \text{ patterns from }M, & \text{if }i>1\\ 
                \emptyset, & \text{Otherwise}
                \end{cases}$
        \State $M=(M-R_{replace})\cup R_{add}$
        \item[\textbf{end for}]
    \end{algorithmic}
\end{algorithm}

The full version of this proposed lifelong learning method can be found \href{https://arxiv.org/abs/1912.01100}{\underline{here}} with an Android App demo for continual object recognition at the edge demo on this \href{https://www.youtube.com/watch?v=Bs3tSjwbHa4&feature=youtu.be}{\underline{YouTube link}}~\cite{pellegrini2019}. 

\begin{figure*}[htbp]
    \centering
    \includegraphics[width=\textwidth]{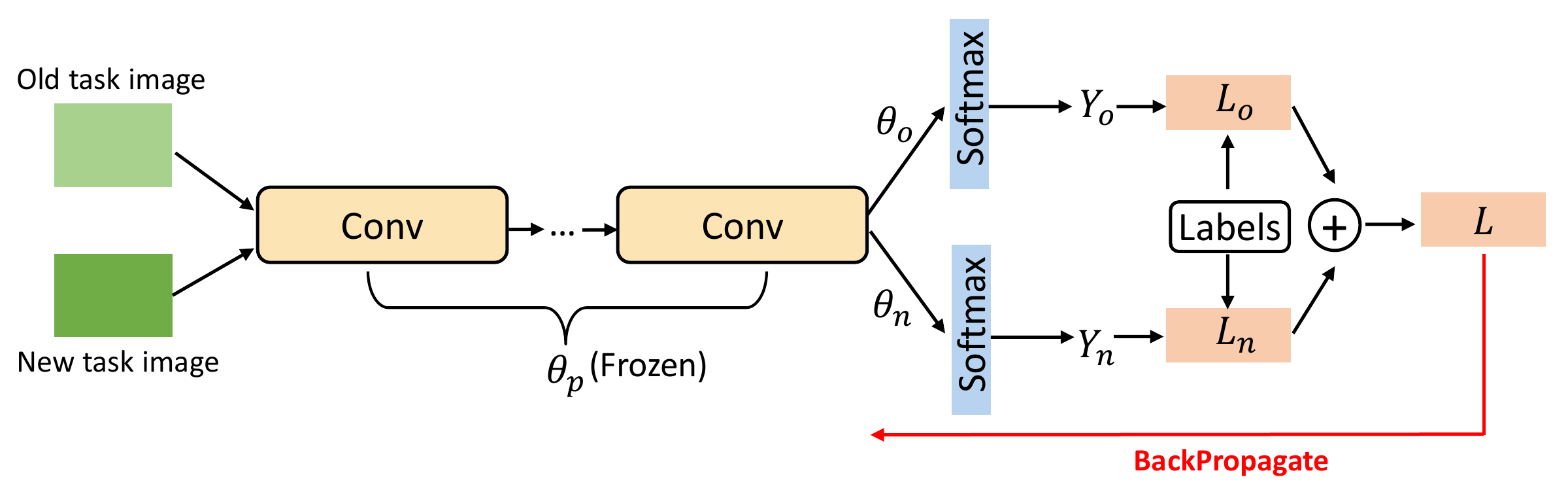}
    \caption{LwF training strategy proposed by Guinness Team.}
    \label{fig:dublin_architecture}
\end{figure*}

\subsection{\textbf{Guinness Team}}
The core backend of the approach was the learning without forgetting (LwF) \cite{Li17learning}. Figure \ref{fig:dublin_architecture} illustrates its training strategy. They deployed a pretrained MobileNet-v2 \cite{sandler2018mobilenetv2}, in which the weights up to the bottleneck are retained as $\theta_p$ ($\theta_p$ here was fine tuned during training) and they trained the bottleneck weights from scratch. Based on LwF, they retained the $\theta_{old}$ that is trained by previous tasks to construct the regularization term for training new weights $\theta_{new}$. It should be noted that there was no replay of previous task images in this structure and only the updated $\theta_{new}$ was retained after training. Empirically, they loaded the initial pretrained weights $\theta_p$ when processing a new task and $\theta_p$ was going to be fine tuned during the training. Details of training scheme are included in Algorithm~\ref{al:training}.

\begin{algorithm}[!ht]
    \caption{Training details}\label{al:training}
    \begin{algorithmic}[1]
        \item[\textbf{Inputs:}]
        \Statex Training images $\mathbf{X}$, labels $\mathbf{Y}$ of the new task and the pretrained parameters $\theta_p$
        \item[\textbf{Initialize:}]
        \Statex $\mathbf{Y}_{o} \gets \mathcal{M}_{\hat{\theta}_p,\theta_{o}}(\mathbf{X})$
               
        \Statex $\theta_{n} \gets$ Xavier-init($\theta_{n}$) 
               
        \Statex Load the pretrained weights $\theta_p$ to the new model
        \item[\textbf{Train:}]
        \Statex $\theta_p^*, \theta_{n}^* \gets {\operatorname*{argmin} \limits_{\hat{\theta}_p,\hat{\theta}_{n}}}(\lambda \mathcal{L}_{o}(\mathbf{Y},\mathbf{Y}_{o})+\mathcal{L}_{n}(\mathbf{Y},\mathbf{Y}_{n}))$ 
        \Statex $\theta_{o} \gets \theta_{n}$ 
    \end{algorithmic}
\end{algorithm}

\subsection{\textbf{Neverforget Team}}
\begin{figure}[htbp]
    \centering
    \includegraphics[width=0.45\textwidth]{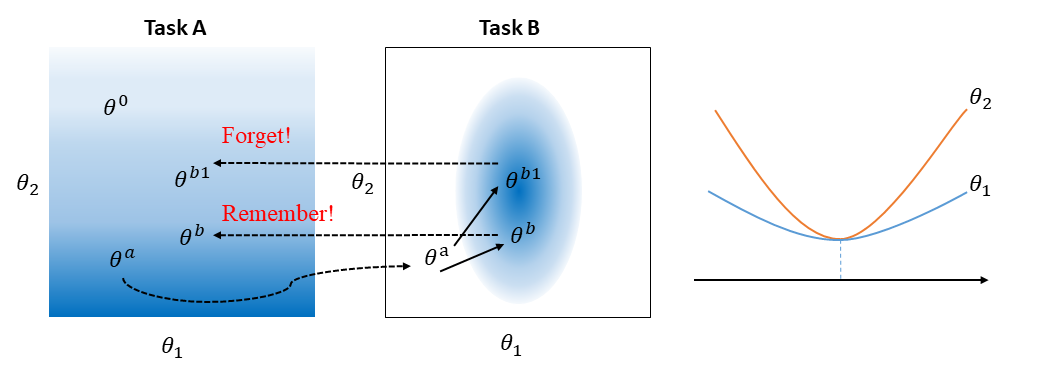}
    \caption{EWC architecture in Neverforget Team's Solution.}
    \label{fig:cuhk_arch}
\end{figure}

The approach was based on Elastic Weight Consolidation (EWC) \cite{Kirkpatrick16}. 
As is shown in the Figure~\ref{fig:cuhk_arch}, the darker area means a smaller loss or a better solution to the task. First, the parameters of the model are initialized as $\theta^0$ and finetuned as $\theta^a$ for Task A. Then, If the model continues to learn Task B and finetuned as $\theta^b1$, the loss of Task A is getting much larger, and it will suffer from the forgetting problem. Instead, the Fisher Information Matrix is utilized to measure the importance of each parameter. If the parameter of the previous task is important, the parameter adjustment in this direction will be constrained and relatively small, if the parameter of the previous task is less important, there will be more space for parameter adjustment in this direction. Assume the importance (the second derivative of log-like function) of parameter $\theta_2$ is more than $\theta_1$, in Task B, the parameter of the neural network will adjust more in $\theta_1$ direction. Thus, the model will gain knowledge of Task B while preserving the knowledge of Task A simultaneously.
The ResNet-101 \cite{he2016deep} was used as the backbone network. The task was sequentially trained on the training set. 
\subsection{\textbf{SDU\_BFA\_PKU Team}}
\begin{figure}[H]
    \centering
    \includegraphics[width=0.45\textwidth]{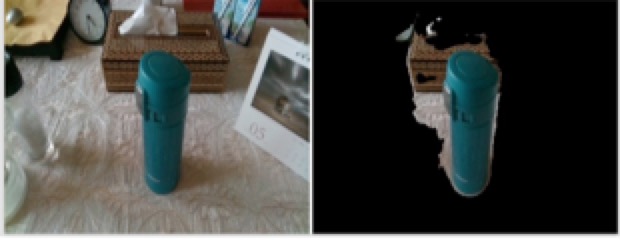}
    \caption{A background removal demo in SDU\_BFA\_PKU Team's solution.}
    \label{fig:pku_loss}
\end{figure}

The approach disentangled this problem with two aspects: background removal problem (See Figure~\ref{fig:pku_loss}) and classification problem.
\begin{figure}[H]
    \centering
    \includegraphics[width=0.45\textwidth]{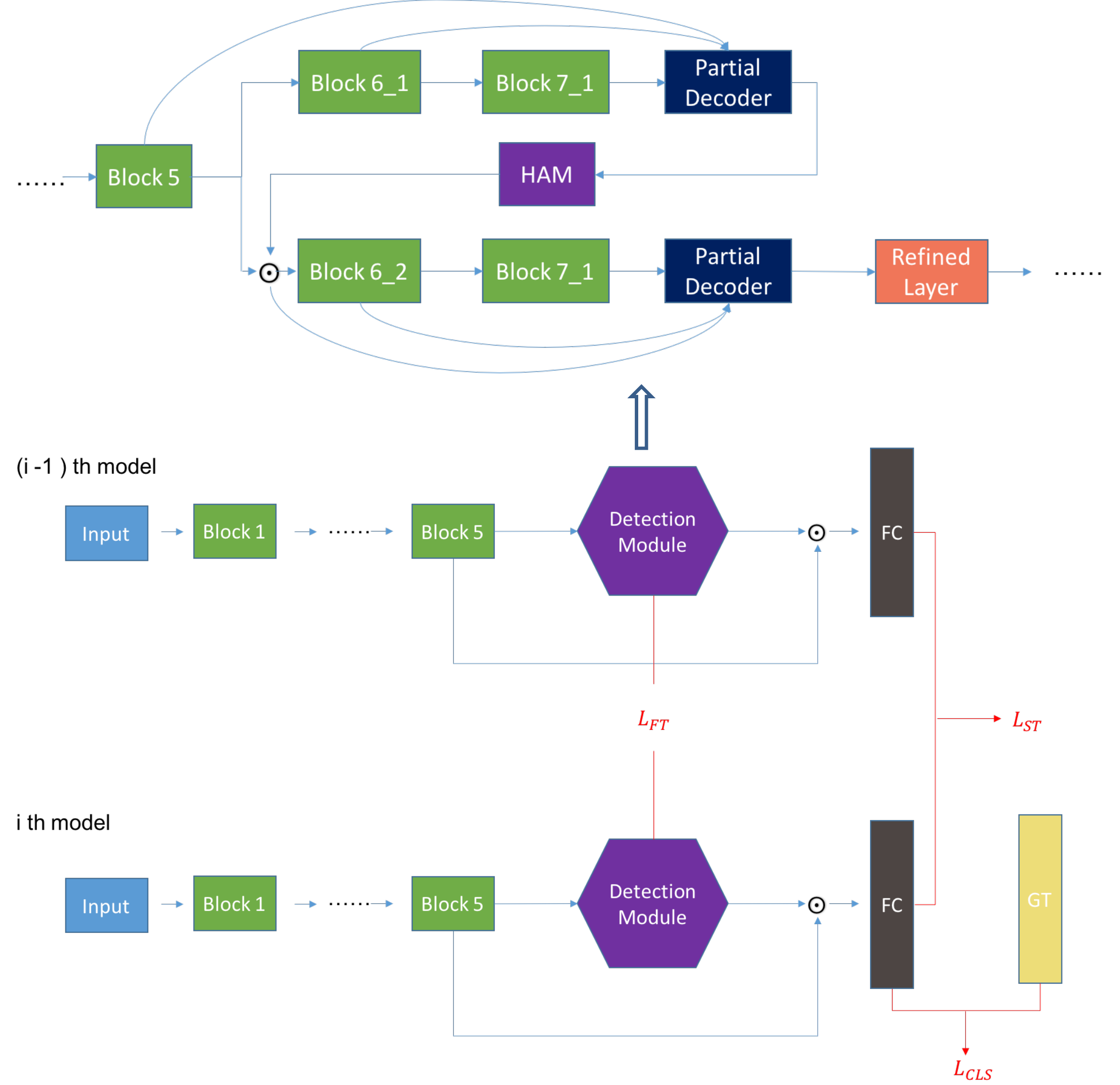}
    \caption{The architecture proposed by SDU\_BFA\_PKU Team.}
    \label{fig:pku_arch}
\end{figure}
First, they utilized saliency detection method to remove the background noise. Cascaded partial decoder framework which
contains two branches is applied to get image saliency map. In each
branch, they used a fast and effective partial decoder. The first
branch generates an initial saliency map which is utilized to
refine the features of the second branch.
For classification problem with catastrophic forgetting, they utilized knowledge distillation to prevent it. They used an auto-encoder as a teacher
translator, and an encoder as student translator, which has
same architecture with teacher translator encoder. The model is aim to project saliency maps from teacher network and student
network to same space. Specifically, For $i$-th task, they regarded
$(i-1)$-th model as teacher network, and $i$-th model as student
network. In order to extract the factor from the teacher network,
they trained the teacher translator in an unsupervised way by
assigning the reconstruction loss at the beginning of every
task training process. Then they utilized student translator to translate
student network's saliency map output, computed $L_{1}$ loss
between teacher network output and student network. In order to save computational and storage size, they used MobileNet-v2 as backbone model~\cite{sandler2018mobilenetv2}. 

\subsection{\textbf{Vidit98 Team}}
This approach sampled validation data from the buffer and use it as replay data. It intelligently creates the replay memory for a task. Here suppose a network is trained on a task $t_n$ and it learns some feature representation of the images in the task, when trained on the task $t_{n+1}$ it learns the feature representation for images in task $t_{n+1}$, but as the distribution of data is task $t_{n+1}$ is different, accuracy drops for images in task $t_{n}$. The replay memory was an efficient representation of previous tasks data whose information was lost. The replay data was sampled from the validation of all the previous tasks. The network on task $t_n$ is trained and the accuracy of batches of validation data is saved. Next, when trained on task $t_i$ ($i > n)$, the accuracy of same batches of validation data of task $t_n$ is calculated. Then they stored the top $k$ batches from validation data of task $t_n$ whose accuracy has dropped the most. This is done for all the tasks $t_{0}$ to $t_{i-1}$. Training for task $t_{i+1}$ they combined the replay data and training data to train for the particular task. The algorithm is shown in Algorithm~\ref{al:iit}. The backbone model they used is MobileNet-v2 \cite{sandler2018mobilenetv2}.
\href{https://github.com/vidit98/Lifelong_Object_Recognition}{\underline{Code}} is made available.  

\begin{algorithm}[!ht]
    \caption{Intelligent resampling method}\label{al:iit}
    \begin{algorithmic}
        \item[\textbf{Results: }Replay\_Data]
        \item[\textbf{Initialization: }]
        \Statex $F_{i}, val\_data_{i}, t_{n}, acc[], best\_acc[], topk$
        \item[\textbf{While} data in $val\_data_{i}$ \textbf{do:}]
        \Statex prec = Accuracy($F_{i}(data)$)
        \Statex Add prec to $acc[]$
        \item[\textbf{end}]
        \item[\textbf{if } $i == n$ \textbf{then:}]
        \Statex Add $acc$ to $best\_acc$
        \item[\textbf{else}]
        \Statex diff = $best\_acc - acc$
        \Statex sort\_diff = sort(diff)
        \Statex Add $topk$ elements corresponding to sort\_diff from $val\_data_{i}$ to Replay\_Data;
    \end{algorithmic}
\end{algorithm}

\subsection{\textbf{HYDRA-DI-ETRI Team}}
The team proposed a selective feature learning method to eliminate irrelevant objects in target images. A Single Shot multibox Detection (SSD) algorithm selected desired objects~\cite{liu2016ssd}.
The SSD algorithm alleviated performance degradation by noisy objects.
Then  SSD weights were trained with annotated images in task $1$, and the refined dataset was fed into a traditional MobileNet~\cite{howard2017mobilenets}.

\begin{figure}[!ht]
\centerline{\includegraphics[scale=0.35]{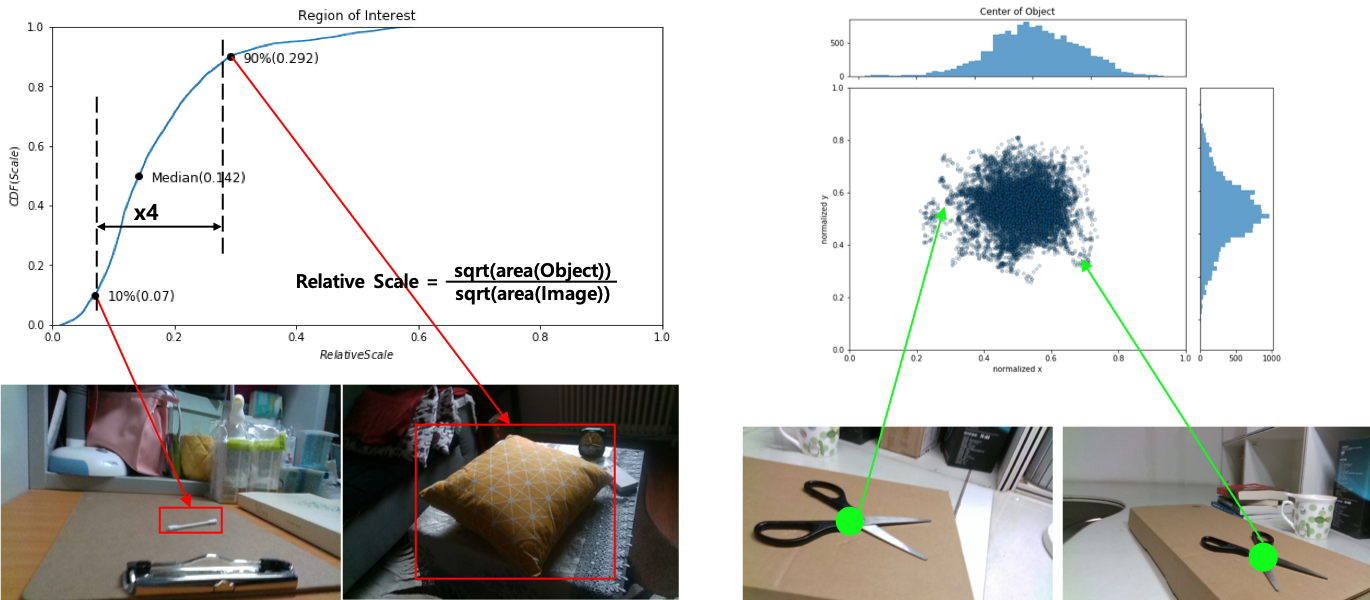}}
\caption{Region of interest analysis in HYDRA-DI-ETRI Team's solution.}
\label{roi_analysis}
\end{figure}

\begin{figure}[!ht]
\centerline{\includegraphics[scale=0.35]{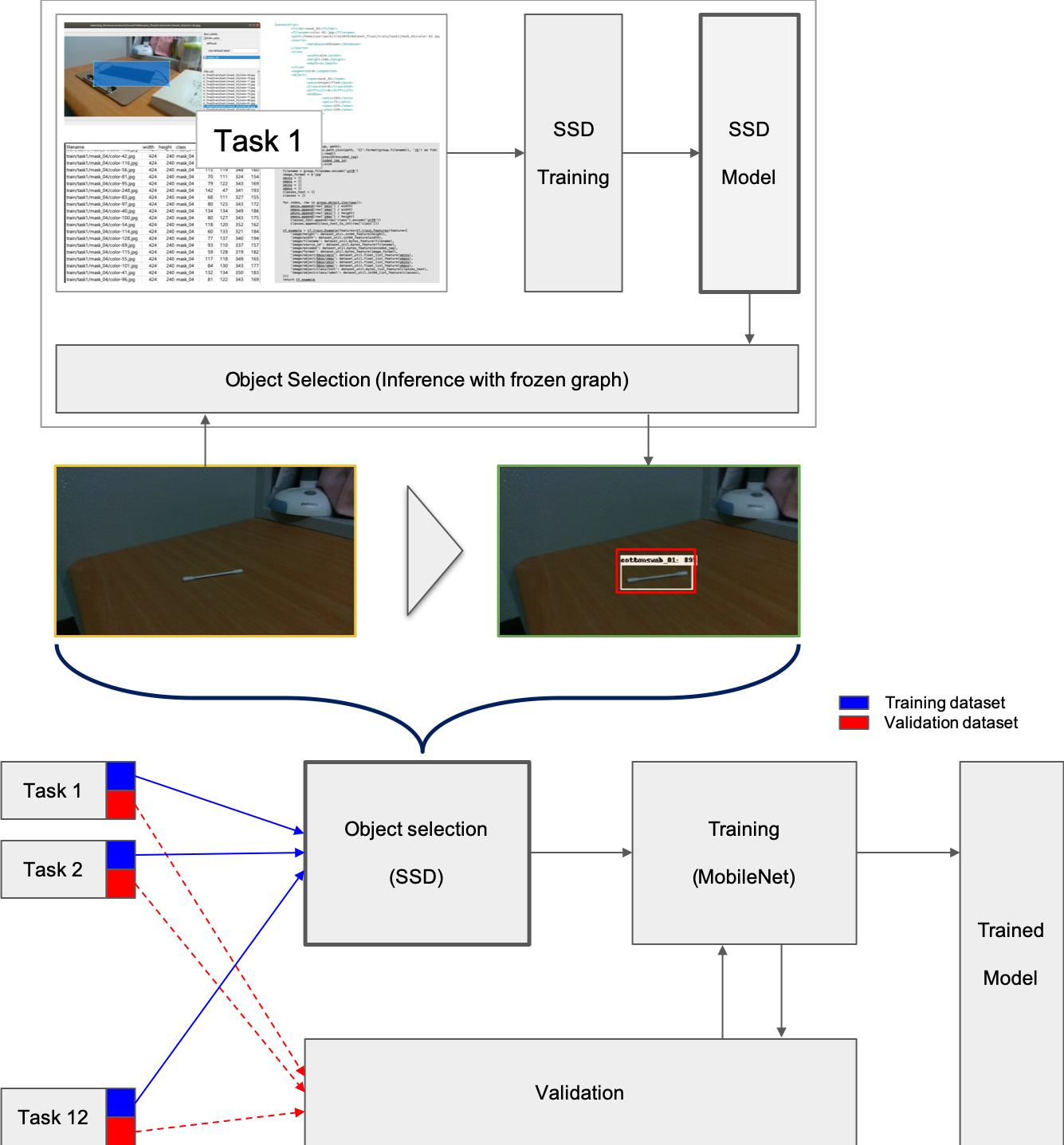}}
\caption{Software architecture for selective feature learning in HYDRA-DI-ETRI Team's solution.}
\label{software_structure}
\end{figure}

The team also analyzed OpenLORIS-Object dataset to design object recognition software (See Figure~\ref{software_structure}), and find that target objects in the dataset coexist with unlabeled objects. The region of interest analysis is illustrated in Figure~\ref{roi_analysis}. Therefore, they proposed a selective feature learning method by eliminating irrelevant features in training dataset. 
The selective learning procedure is as follows:
1) extracting target objects from training dataset by an object detection algorithm, 
2) feeding the refined dataset into a deep neural network to predict labels.
In their software, they applied to a SSD as the object detection algorithm due to convenience of flexible feature network design and proper detection performances.

\subsection{\textbf{NTU\_LL Team}}
\begin{figure*}[htbp]
    \centering
    \includegraphics[width=\textwidth]{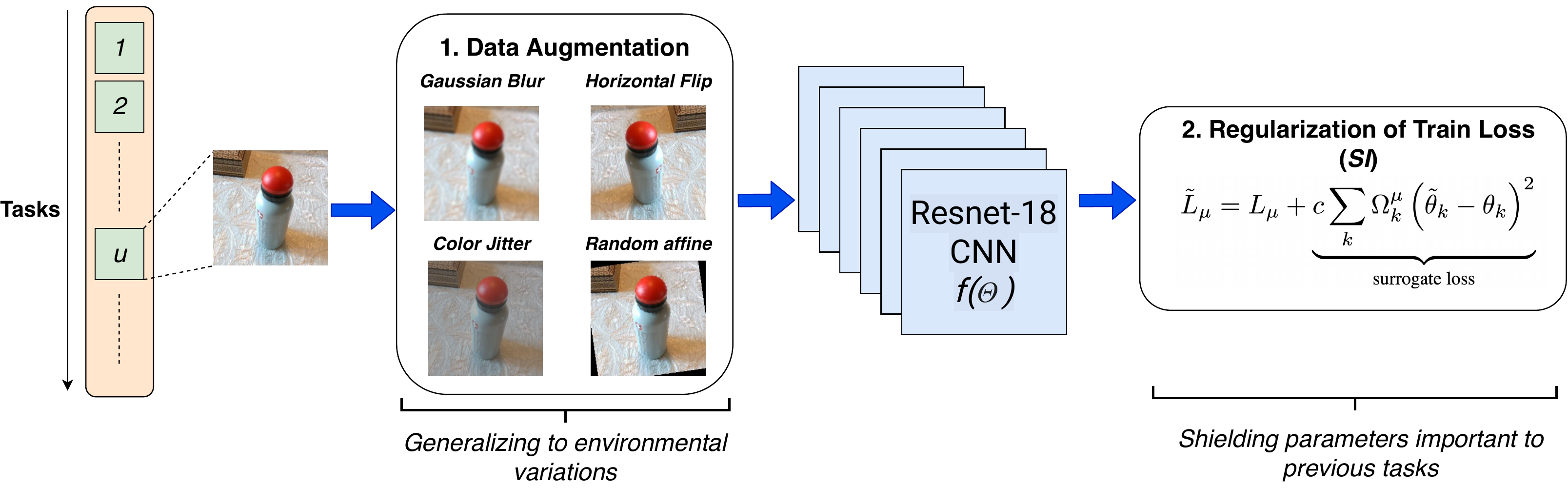}
    \caption{Solution architecture proposed by NTU\_LL Team.}
    \label{fig:ntu_arch}
\end{figure*}

The team utilized a combination of Synaptic
Intelligence (SI) based regularization method and data augmentation~\cite{zenke2017continual} (See Figure~\ref{fig:ntu_arch}). The augmentation strategies they applied were Color Jitter and Blur. ResNet-18 was used for backbone model~\cite{he2016deep}. 
\newpage
\section{Finalists Information}
\noindent\textbf{HIK\_ILG Team}\\
\textbf{\textit{Title: Dynamic Neural Network for Incremental Learning}}\\
\textbf{\textit{Members: }}Liang Ma$^{1}$, Jianwen Wu$^{1}$, Qiaoyong Zhong$^{1}$, Di Xie$^{1}$ and Shiliang Pu$^{1}$\\
\textbf{\textit{Affiliation: }}$^{1}$ Hikvision Research Institute, Hangzhou, China.\\

\noindent\textbf{Unibo Team}\\
\textbf{\textit{Title: Efficient Continual Learning with Latent Rehearsal}}\\
\textbf{\textit{Members: }}Gabriele Graffieti$^{1}$, Lorenzo Pellegrini$^{1}$, Vincenzo Lomonaco$^{1}$ and Davide Maltoni$^{1}$\\
\textbf{\textit{Affiliation: }}$^{1}$University of Bologna, Bologna, Italy.\\

\noindent\textbf{Guinness Team}\\
\textbf{\textit{Title: Learning Without Forgetting Approaches for Lifelong Robotic Vision}}\\
\textbf{\textit{Members: }}Zhengwei Wang$^{1}$, Eoin Brophy$^{2}$ and Tom\'as E. Ward$^{2}$\\
\textbf{\textit{Affiliation: }}$^{1}$Zhengwei Wang is with V-SENSE, School of Computer Science and Statistics,
        Trinity College Dublin, Dublin, Irleand; $^{2}$Eoin Brophy and Tom\'as E. Ward are with the Inisht Centre for Data Analytics, School of Computing, Dublin City University, Dublin, Ireland.\\
        
\noindent\textbf{Neverforget Team}\\
\textbf{\textit{Title: A Small Step to Remember: Study of Single Model VS Dynamic Model}}\\
\textbf{\textit{Members: }}Liguang Zhou$^{1,2}$\\
\textbf{\textit{Affiliation: }}$^{1}$The Chinese University of Hong Kong (Shenzhen),Shenzhen, China, $^{2}$Shenzhen Institute of Artificial Intelligence and Robotics for Society, China.\\

\noindent\textbf{SDU\_BFA\_PKU Team}\\
\textbf{\textit{Title: SDKD: Saliency Detection with Knowledge Distillation}}\\
\textbf{\textit{Members: }}Lin Yang$^{1,2,3}$\\
\textbf{\textit{Affiliation: }}$^{1}$Peking University, Beijing, China, $^{2}$Shandong University, Qingdao, China,  $^{3}$Beijing Film Academy, Beijing, China.\\

\noindent\textbf{Vidit98 Team}\\
\textbf{\textit{Title: Intelligent Replay Sampling for Lifelong Object Recognition}}\\
\textbf{\textit{Members: }}Vidit Goel$^{1}$, Debdoot Sheet$^{1}$ and Somesh Kumar$^{1}$\\
\textbf{\textit{Affiliation: }}$^{1}$Indian Institute of Technology, Kharagpur, India.\\

\noindent\textbf{HYDRA-DI-ETRI Team}\\
\textbf{\textit{Title: Selective Feature Learning with Filtering Out Noisy Objects in Background Images}}\\
\textbf{\textit{Members: }}Soonyong Song$^{1}$, Heechul Bae$^{1}$, Hyonyoung Han$^{1}$ and Youngsung Son$^{1}$\\
\textbf{\textit{Affiliation: }}$^{1}$Electronics and Telecommunications Research Institute (ETRI), Korea.\\

\noindent\textbf{NTU\_LL Team}\\
\textbf{\textit{Title: Lifelong Learning with Regularization and Data Augmentation}}\\
\textbf{\textit{Members: }}Duvindu Piyasena$^{1}$, Sathursan Kanagarajah$^{1}$, Siew-Kei Lam$^{1}$ and Meiqing Wu$^{1}$\\
\textbf{\textit{Affiliation: }}$^{1}$Nanyang Technological University, Singapore.\\

\section*{ACKNOWLEDGMENT}
The work was partially supported by a grant from the Research Grants Council of the Hong Kong Special Administrative Region, China (Project No. CityU 11215618).
The authors would like to thank Hong Pong Ho from Intel RealSense Team for the technical support of RealSense cameras for recording the high-quality RGB-D data sequences.

\bibliography{root}
\bibliographystyle{IEEEtran}

\end{document}